\documentclass[10pt, a4paper]{article}
\usepackage{lrec2022} 
\usepackage{multibib}
\newcites{languageresource}{Language Resources}
\usepackage{graphicx}
\usepackage{tabularx}
\usepackage{soul}
\usepackage{titlesec}
\titleformat{\section}{\normalfont\large\bfseries\center}{\thesection.}{1em}{}
\titleformat{\subsection}{\normalfont\SmallTitleFont\bfseries\raggedright}{\thesubsection.}{1em}{}
\titleformat{\subsubsection}{\normalfont\normalsize\bfseries\raggedright}{\thesubsubsection.}{1em}{}
\renewcommand\thesection{\arabic{section}}
\renewcommand\thesubsection{\thesection.\arabic{subsection}}
\renewcommand\thesubsubsection{\thesubsection.\arabic{subsubsection}}

\usepackage{epstopdf}
\usepackage[utf8]{inputenc}

\usepackage{hyperref}
\usepackage{xstring}

\usepackage{color}

\title{Analysis of Co-Laughter Gesture Relationship on RGB videos in Dyadic Conversation Context}

\name{Hugo Bohy\textsubscript{1}, Ahmad Hammoudeh\textsubscript{123}, Antoine Maiorca\textsubscript{1}, Stéphane Dupont\textsubscript{2}, Thierry Dutoit\textsubscript{13}} 

\address{\textsuperscript{1} ISIA Lab, \textsuperscript{2} MAIA Lab,\textsuperscript{3} TRAIL \\
\textsuperscript{1,2} University of Mons, Mons, Belgium, \textsuperscript{3} Wallonia-Brussels Federation, Belgium \\
\{hugo.bohy, 535653, antoine.maiorca, stephane.dupont, thierry.dutoit\}@umons.ac.be\\}

\abstract{
The development of virtual agents has enabled human-avatar interactions to become increasingly rich and varied. Moreover, an expressive virtual agent i.e. that mimics the natural expression of emotions, enhances social interaction between a user (human) and an agent (intelligent machine). The set of non-verbal behaviors of a virtual character is, therefore, an important component in the context of human-machine interaction. Laughter is not just an audio signal, but an intrinsic relationship of multimodal non-verbal communication, in addition to audio, it includes facial expressions and body movements. Motion analysis often relies on a relevant motion capture dataset, but the main issue is that the acquisition of such a dataset is expensive and time-consuming. This work studies the relationship between laughter and body movements in dyadic conversations. The body movements were extracted from videos using deep learning based pose estimator model. We found that, in the explored \textit{NDC-ME} dataset, a single statistical feature (i.e, the maximum value, or the maximum of Fourier transform) of a joint movement weakly correlates with laughter intensity by 30\%. However, we did not find a direct correlation between audio features and body movements. We discuss about the challenges to use such dataset for the audio-driven co-laughter motion synthesis task.
\\ \newline 
}

\begin{document}

\maketitleabstract

\section{Introduction}
\label{sec:intro}
The interactive gesture generation task aims to control the gesture of a virtual character with a user control signal. Many works addressed the problem of synthesizing the gesture of an avatar along with a speech modality \cite{alexanderson2020style,ahuja-etal-2020-gestures}. These methods enabled capturing and synthesis of natural co-speech gestures of a virtual character. \cite{kucherenko2020gesticulator} used speech and text jointly as inputs to their proposed model to generate the gestures and reported that the multimodal aspect of their method helps to understand the sentence semantics and outputs natural and diverse gestures. \cite{Yoon2020Speech} encoded these modalities along with the speaker identity since each expressive behavior highly relies on the speaker.
\par \noindent
\\
Nevertheless, motion synthesis from a non-verbal audio input such as laughter is a complex task where no a priori semantic information is available with the audio signal to help with understanding the overall context. However, laughter constitutes an important part of social interaction \cite{mckeown2015relationship} where the smiling and laughing expression of an interlocutor induces a mimicry effect on each partner \cite{el2019smile}. The growing interest in virtual environments has led to the development of virtual social agents. The immersive factor of a virtual world is partly induced by the naturalness of the motion of virtual characters. The human-avatar social interaction is an active research topic among the computer vision community and rendering natural motion is a crucial task to enhance the social aspect of the avatar \cite{garau2003impact}. Co-laughter gesture synthesis is thus a relevant task in human computer interaction where it can be exploited in various use cases such as video game development \cite{mancini2013laugh} or in a medical context e.g. to enhance the social skills of children with autism spectrum disorder \cite{didehbani2016virtual}. 
 \par \noindent
 \\
 The work presented in this paper falls in a wider project aiming at generating co-laughter motion corresponding to the audio given at its input using generative deep neural networks. We present here first analyses results on the relationship between body movements (excluding facial expressions) and several aspects of laughter. These analyses would help us gain a better understanding of our data and thus organize their use to build the previously mentioned generative system. 
The motion data is not extracted from motion capture sensors but is estimated from the recorded RGB videos directly. Neural networks are powerful tools for learning complex relationships between given modalities within a database. Thus, the proposed analysis allows us to identify whether correlations between laughter, its intensity and the associated movement are significant within a given dataset. If this dataset does not exhibit a high correlation between laughter and body motion, it may be a challenging dataset to train neural networks that synthesize body motion from audio laughter.
\par \noindent
\\
This paper is organized as follow: Section \ref{sec:sota} reviews the state-of-the-art analysis of the relationship between multiple laughter modalities and co-laughter motion synthesis methods. Section \ref{sec:exp} explains the experimental protocol and Section \ref{sec:res} analyzes the experimental results. Section \ref{sec:futur} discusses the limitations of this work and proposes some improvements.

\section{Related Work}
\label{sec:sota}
To focus on the synthesis task, it is useful to understand and measure the relationship between laughter as an audio signal and the gesture performed during that laughter. \cite{6681455} found a significant contrast in the captured motions between different types of laughter (hilarious, social, and non-laughter) and claimed that motion features analysis helped with the classification of laughter type. \cite{7298420} showed that full-body motion features are sufficient to detect laughter occurrences. \cite{mancini2013laugh} pointed out the periodic pattern of the shoulder motion while laughing in the dataset \textit{Multimodal Multiperson Corpus of Laughter in Interaction} \cite{10.1007/978-3-319-02714-2_16}. \cite{ishi2019analysis} focused on laughter intensity to reveal that the degree of smiling face and the occurrences of the front, back, up, and down motions are proportional to the laughter intensity.
\par \noindent
\\
\cite{dilorenzo2008laughing} proposes a physics-based model to synthesize the torso deformation induces by the air flow while laughing. \cite{niewiadomski2014rhythmic} performs a harmonic analysis of the laughter body motions to get relevant rhythmic features for the generation of body movements. \cite{ding2017audio} synthesized upper body gestures from laughter audio signal based on the captured or defined co-laughter motion correlations. Their approach is based on a statistical framework for head and torso motion and a rule-based method for shoulder motion due to the limitation of their dataset. \cite{ishi2019analysis} generated co-speech and laughter motion (eyelids, face, hand and upper body) on physical android robots. The works presented above relied on recorded motion capture datasets of people laughing in multiple contexts. \cite{jokinen2016body} analyzed  videos of social interactions and pointed out the synchrony of body movements with laughter. Similarly, this research aims to identify body motion relationships with laughter from RGB videos and audio signals. However, \cite{jokinen2016body} estimated bounding boxes around the limbs of the participants. 
\par \noindent
\\
This work proposes an analysis of the relationship between low-level motion features extracted from RGB videos i.e. the Cartesian position of each joint, the laughter intensity and audio features in the context of a dyadic conversation. This relational study aims to identify any significant correlation between the positions of the joints and the laughter audio signal and intensity. Two approaches are tested and are further explained in Section \ref{subsubsec:audio} regarding the laughter audio signal: first, the audio signal is decomposed into a set of low-level and physical features and then the audio signals are embedded into a latent space from the baseline speech oriented model \textit{Wav2vec 2.0} \cite{DBLP:journals/corr/abs-2006-11477}.
Finally, the relationship between the 2D Cartesian positions of the skeleton and laughter intensity is established and described in Section \ref{subsubsec:intensity}.
\section{Experiments}
 \label{sec:exp}
\subsection{Dataset}
\begin{figure}
    \centering
    \includegraphics[scale=0.15]{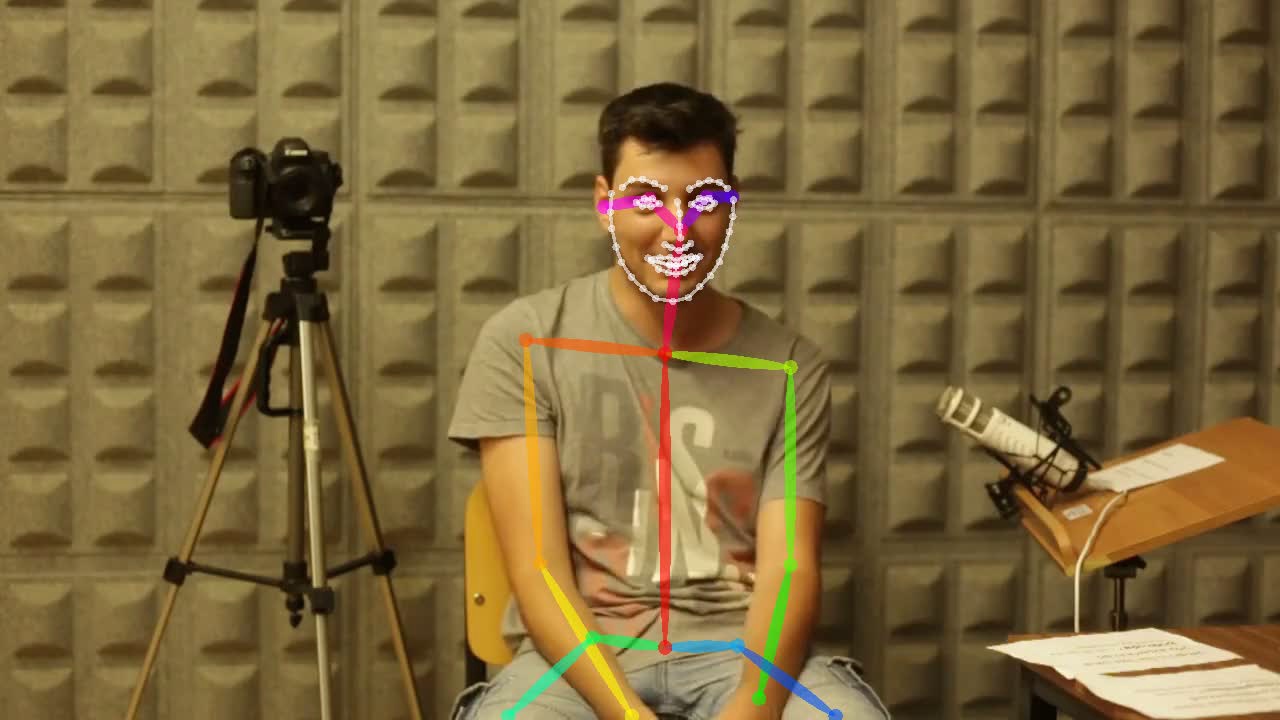}
    \includegraphics[scale=0.25]{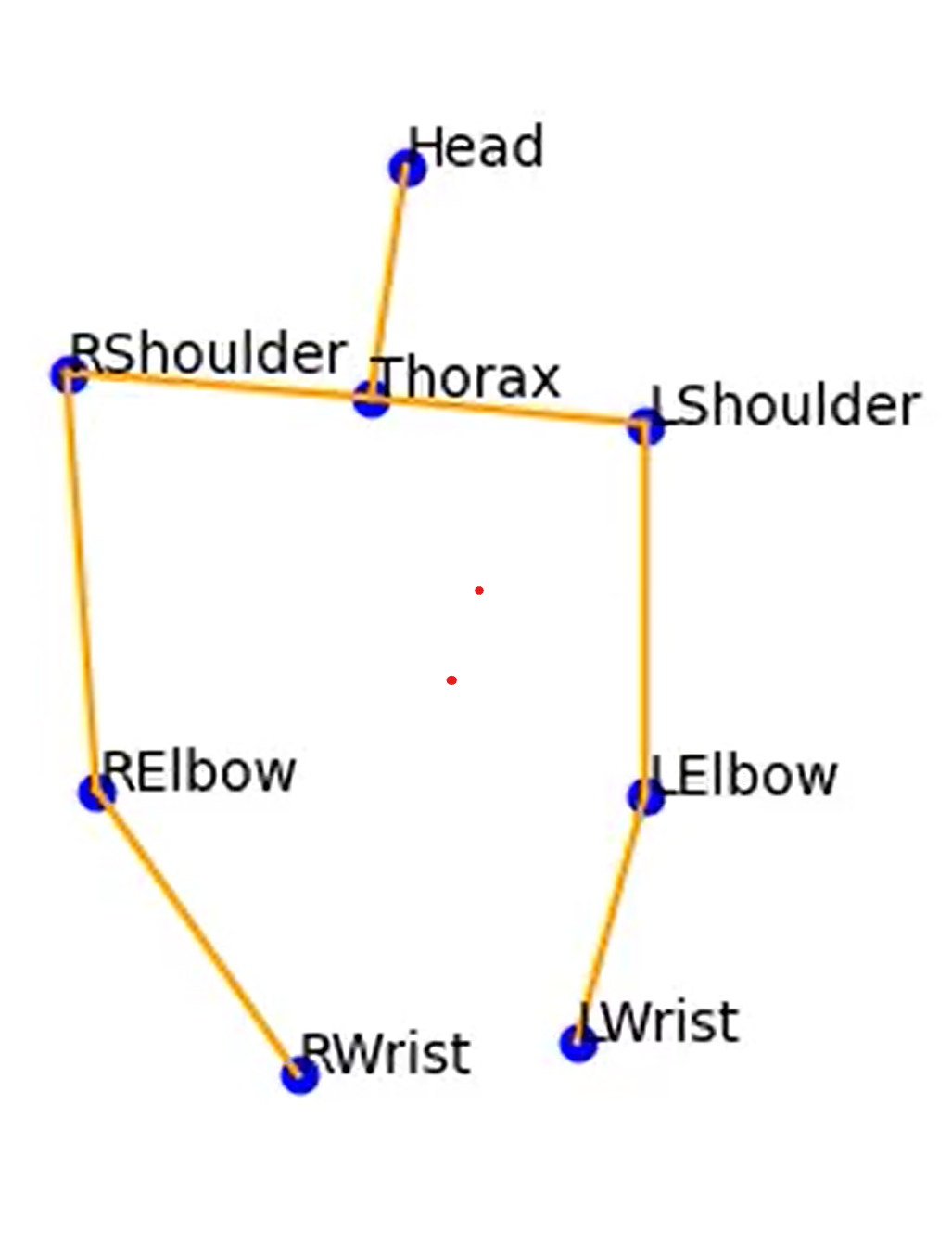}
    \caption{Top: sample of a video with the estimated skeleton and face landmarks. Since this work only focuses on the body skeleton, the face landmarks are ignored. Bottom: structure of the upper body skeleton.}
    \label{fig:openposed}
\end{figure}

In our experiments, we used the dataset \textit{Naturalistic Dyadic Conversation on Moral Emotions} (\textit{NDC-ME}) \cite{heron2018dyadic}. It consists of a collection of dyadic conversations focusing on moral emotions through speaker-listener interactions. In contrast to \textit{IFADV} Corpus \cite{van2008ifadv} and the \textit{Cardiff Conversation Database} \cite{aubrey2013cardiff}, the whole upper body of the participants is available in the videos and their motion is not constrained by any object. 21 pairs of participants have been recorded while they were interacting together without following a fixed scenario. The audio and videos have been captured separately. The emotions and the intensity of the expressed emotion of each participant during the recording have been labeled using the annotation tool \textit{ELAN} \cite{elan} and are available here \footnote{\href{https://zenodo.org/record/3820510}{https://zenodo.org/record/3820510}}. The annotation rules follow the protocol \footnote{This protocol is available \href{https://www.researchgate.net/publication/341371010_supplementary_materialpdf}{here}} used by \cite{el2019smile}. The laughter clips are also labeled into 3 categories regarding their intensity: low, medium, and high. At that time, only 7 pairs have been annotated. Following these annotations, the audio and videos in which laughter occurs are extracted from the initial dataset. 186 videos are kept including 10 male and 4 female speakers for a total duration of 199.33 seconds. Then, 2D Cartesian positions of the skeleton joints are extracted from the RGB videos using \textit{OpenPose} \cite{DBLP:journals/corr/abs-1812-08008}. The skeleton consists of 8 joints representing the upper body of the subject. A frame sample with an estimated skeleton as well as the upper body structure is shown in Figure \ref{fig:openposed}. 

\subsection{Experimental setup}
\label{sec:exo}
This part describes the experimental protocol to identify the correlation between the laughter modalities in \textit{NDC-ME} dataset.
\par \noindent
\\
Joint movement signals are represented as time series $s$ where $s_{j}^{i} = p_{j}^{i} - \bar{p_{j}}$ with $ p_{j}^{i} $, the Cartesian position of a joint $j$ at frame $i$ and $ \bar{p_{j}}$ the mean position of the joint $j$. Thus, $s_j$ is the temporal fluctuations of the position of the joint $j$ around its mean position. Then, the horizontal and vertical component of the motion signal of joint $j$ are respectively noted $x_j$ and $y_j$. In this work, we consider separately horizontal and vertical movements for the sake of simplicity but it would be interesting to consider both directions. The correlations on shoulders, elbows and wrists are computed separately for the right and left body parts and we further report the average value.

\subsubsection{Body movement and audio features}
\label{subsubsec:audio}
We wanted to analyse the correlation between the audio signal and the body movement. For the audio signal, we extracted two sets of features per 20 ms frame : one that includes 19 well-known low-level features in the speech analysis domain (3 from LPC, 13 MFCCs and 3 LPCCs), and the other that includes the 512 embedded outputs of the \textit{Wav2vec 2.0} model. For each subset of features, we computed the pearson correlation coefficient between $(x_j,y_j)$ and the time series of audio features.


\subsubsection{Body movements and laughter intensity}
\label{subsubsec:intensity}
Firstly, the following features were extracted for each horizontal and vertical joint movement signal $(x_j,y_j)$: In the time domain (power $P$, maximum amplitude value $max$, mean value $\mu$  and standard deviation $\sigma$), and the frequency domain (the maximum value of Fourier Transform $max(FT)$, the mean of Fourier Transform $\mu(FT)$, and peak frequency $f_\mathit{pk}=argmax(FT)$). Since laughter videos vary in length, Fourier Transform curves were linearly interpolated in 248 uniform samples between 0 and Nyquist frequency $f_\mathit{Nyquist}$. The upper 10\% of the frequency range was excluded when finding the peak frequency in order to exclude high-frequency noise ( $f_\mathit{pk} < 0.9 f_\mathit{Nyquist}$ ). The correlation between those extracted features of joints movement and laughter intensity are then analyzed.
\begin{table*}
\centering
\begin{center}
\begin{tabular}{ | p{1.1cm} || p{6.55cm} || p{6.55cm} |}
\hline
\textbf{Feature} &
\centering \textbf{Horizontal Movement}& 
\centering \textbf{Vertical Movement}
\end{tabular}
\begin{tabular}{ | p{1.1cm} || p{0.8cm} | p{1.0cm} | p{1.3cm} | p{0.9cm}| p{0.8cm}||  p{0.8cm} | p{1.0cm} | p{1.3cm} | p{0.9cm}| p{0.8cm}|}
\hline
\textbf{} &
\textbf{Head} &
\textbf{Thorax} &
\textbf{Shoulders} &
\textbf{Elbows} &
\textbf{Wrists} &
\textbf{Head} &
\textbf{Thorax} &
\textbf{Shoulders} &
\textbf{Elbows} &
\textbf{Wrists} 
\\
\hline
LPC &
 0.03&   0.02&  0.05&  0.03 & 0.02  
& 0.04 & 0.05  &  0.07 &  0.02 & 0.03  \\
\hline
MFCCs &
 -0.03&   -0.01&  0.01&  -0.01 & 0.01  
& -0.08 & -0.06  &  -0.06 &  -0.04 & -0.01  \\
\hline
LPCCs &
 0.05&   0.03&  0.04& -0.01 & -0.01  
& 0.05 & 0.07  &  0.07 &  -0.02 & 0.08  \\
\hline
W2V&
 \textbf{0.09}&   \textbf{0.08}&  \textbf{0.07}&  \textbf{0.08} & \textbf{0.09}  
& \textbf{0.11} & \textbf{0.09}  & \textbf{0.09} &  \textbf{0.10} & \textbf{0.09}   \\
\hline

\end{tabular}
\caption{Maximum average correlation between an audio feature and a joint with respect to its movement direction.}
\label{table:table6}
\end{center}
\end{table*}

\begin{table*}
\centering
\begin{center}
\begin{tabular}{ | p{1.1cm} || p{6.55cm} || p{6.55cm} |}

\hline
\textbf{Feature} &
\centering \textbf{Horizontal Movement}& 
\centering \textbf{Vertical Movement}
\end{tabular}

\begin{tabular}{ | p{1.1cm} || p{0.8cm} | p{1.0cm} | p{1.3cm} | p{0.9cm}| p{0.8cm}||  p{0.8cm} | p{1.0cm} | p{1.3cm} | p{0.9cm}| p{0.8cm}|}
\hline
\textbf{} &
\textbf{Head} &
\textbf{Thorax} &
\textbf{Shoulders} &
\textbf{Elbows} &
\textbf{Wrists} &
\textbf{Head} &
\textbf{Thorax} &
\textbf{Shoulders} &
\textbf{Elbows} &
\textbf{Wrists} 
\\
\hline
\hline
max &
0.09&  0.25&  0.30&  0.22&  0.26
& 0.26&  \textbf{0.39}&  \textbf{0.25}&  0.25&  0.20 \\
\hline

P &
0.08&  0.09&  0.18&  0.10&  0.13 
& 0.29&  0.25&  0.16&  0.10 &  0.10 \\
\hline

$\mu$ &
0.10&  0.05&  0.06&  0.02&  0.08
& -0.17&  -0.19&  -0.14&  -0.16&  -0.15  \\
\hline
$\sigma$ &
0.16&  0.23&  0.28&  0.23&  0.26
& 0.35&  0.31&  0.21&  0.27&  0.20  \\
\hline

$\mu(FT)$ &
0.13&  0.26&  0.30&  0.24&  0.26
& 0.28& 0.37&  0.23&  0.25&  0.18 \\
\hline

max(FT) &
0.23&  \textbf{0.32}&  \textbf{0.36}&  \textbf{0.32}&  \textbf{0.34}
& \textbf{0.36}&  0.33&  0.24&  \textbf{0.32}&  \textbf{0.21} \\
\hline

fpk &
\textbf{-0.29}&  -0.22&  -0.20&  -0.2&  -0.22
& -0.22&  -0.21&  -0.12&  -0.20&  -0.12  \\
\hline

\end{tabular}
\caption{Correlation between laughter intensity and a joint movement feature. The power $P$, maximum amplitude value $max$, mean value $\mu$  and standard deviation $\sigma$ are computed from the horizontal and vertical motion signals in the time domain. In the frequency domain, the motion features  are the maximum value of Fourier Transform $max(FT)$, the mean of Fourier Transform $\mu(FT)$ and and peak frequency $f_{pk}$. The correlation is bound between -1 and 1. The higher absolute value means a stronger correlation and 0 shows no correlation in the data.}
\label{table:table5e}
\end{center}
\end{table*}

\section{Results}
\label{sec:res}
Section \ref{sec:res} presents the results of the correlation analysis between body movements, audio features and laughter intensity.

\subsection{Body movements and audio features}
Table \ref{table:table6} shows the maximum average correlation between an audio feature and a joint movement. The values depicted informs us about the weak correlation between the evolution of the position of a joint compared to the evolution of an audio feature. However, using embedded features rather than interpretable ones increases the correlation across all joints.

\subsection{Body movements and laughter intensity}
\begin{figure}[t]
    \centering
    \includegraphics[scale=0.45]{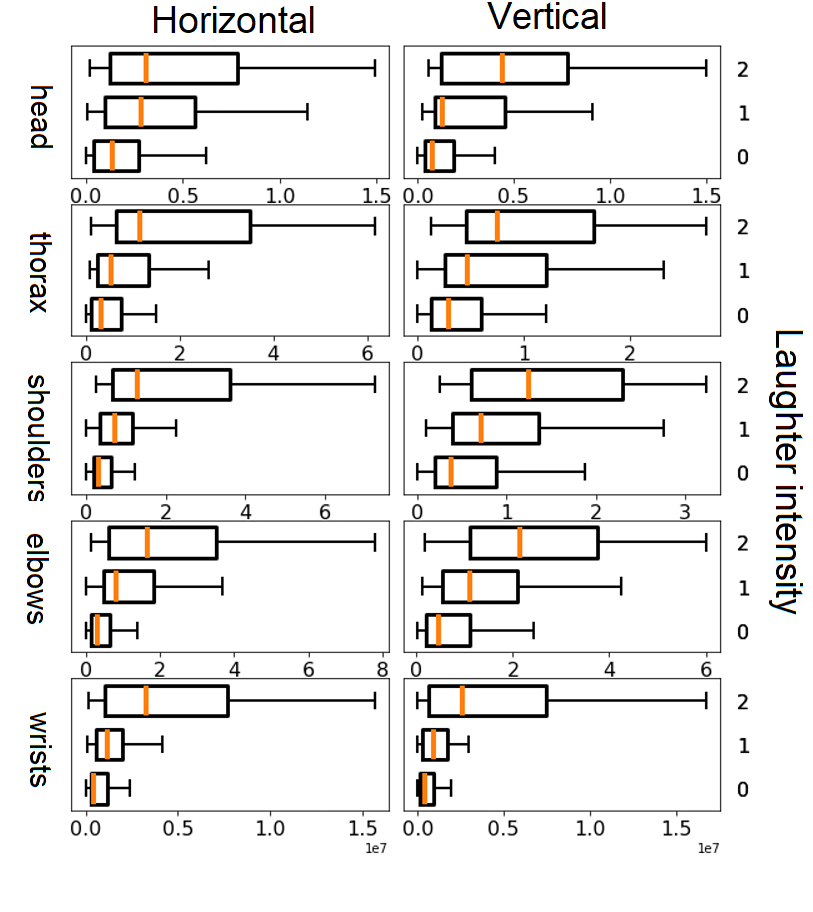}
    \caption{The maximum Fourier transform of a joint movement signal $max(FT(p_j))$ under multiple laughter intensities. Each Row represents a joint and each column represents a direction of movement (horizontal/vertical). Each figure has 3 boxplots (low laughter intensity at 0, medium at 1, and high at 2). The orange line in a boxplot represents the mean.}
    \label{fig:maxft}
\end{figure}

The correlation between the extracted features and laughter intensity is shown in table \ref{table:table5e}. Since $max(FT)$ feature has the highest correlation, we visualized the distribution of $max(FT)$ features under multiple laughter intensities in Figure \ref{fig:maxft}. The visualization of $max(FT)$, similar to the other extracted features, resulted in overlapping boxplots. Hence, we conclude that any of the extracted features alone is not sufficient to identify the laughter intensity. However, statistically speaking, the mean value of the distribution (the orange line in Figure \ref{fig:maxft}) increases with laughter intensity.

\section{Discussion and Challenges}
\label{sec:futur} 

The results presented in Section \ref{sec:res} indicate that, in \textit{NDC-ME} dataset, body movements and audio features seem to be weakly correlated. Further investigation and processing are needed to draw a more robust conclusions. Thus, this dataset seems, at the moment and with this current analysis, challenging for a co-laughter gesture synthesis task. However, we found some aspects in the dataset that might impact the results in our analysis: in some files, speaker speech overlaps with the listener's laughter and we suspect that this influenced the experimental results in Section \ref{sec:res}. These need to be removed from the dataset in future work to get more accurate results. One suggestion is the application of channel source separation methods to the audio to distinguish the laughter or speech of each participant and have a better audio representation (more suitable features). Then, the laughter intensity has been subjectively annotated by a single annotator and having a low number of annotators makes the data distribution more sensitive to human error. We suggest to increase the number of annotators and e.g. extracting the mean annotations to reduce this impact. Moreover, since the dataset has not been fully annotated yet, it contains a relatively small amount of laughter examples. Then, in a future work, we would like to extract correlations from audio acoustic features such as pitch or loudness. Moreover, it would be interesting to take into account other modalities such as the type of laughter and the context of the interactions. Finally, in this work, we focus on body movement but face landmarks are available from the \textit{OpenPose} estimation as shown in Figure \ref{fig:openposed}. The relationship between those landmarks and the laughter intensity and laughter audio features can be established in further investigation.

\section{Conclusion}
\label{sec:ccl}
This work proposes a method to analyze the relationship between laughter, its intensity and the body movement in recorded dyadic conversations. In contrast with previous works, the gestures are extracted from the RGB videos using a baseline pose estimation method. First, this work highlights around 30\% correlation between laughter intensity and motion features where the maximum amplitude of the Fourier transform leads to the highest correlation value. Moreover, the analysis of correlation between interpretable and high-level audio features does not output significant correlation values. This work highlights some of the limitations of \textit{NDC-ME} dataset that we need to take into account in the context of deep generative model training for body motion generation from a laughter audio signal. This analysis opens the way to create datasets suited to build multimodal models that generate the motion of virtual agents from the audio cue.

\section{Acknowledgements}
This work was supported by Service Public de Wallonie Recherche under grant n° 2010235 - \textit{ARIAC} by \textit{DIGITALWALLONIA4.AI}
\section{Bibliographical References}\label{reference}

\bibliographystyle{lrec2022-bib}
\bibliography{lrec2022-example}

\end{document}